\documentclass[journal]{IEEEtran}
\ifCLASSINFOpdf
\else
\fi
\usepackage{graphicx}
\usepackage{amsmath}  
\usepackage{mathtools}
\usepackage{algorithm}
\usepackage[noend]{algpseudocode}
\usepackage{booktabs}
\usepackage{subcaption}
\usepackage{multicol}
\usepackage{framed} 
\usepackage[framed]{ntheorem}
\newframedtheorem{frm-thm}{Premise}
\newframedtheorem{frm-lmm}{Lemma}
\newframedtheorem{frm-rule}{``Condensed'' Rule}
\usepackage{amssymb}
\usepackage{balance}

\begin{document}
%
\title{A Visual Communication Map for Multi-Agent Deep Reinforcement Learning}
%
%
%

\author{Ngoc~Duy~Nguyen,
        Thanh~Thi~Nguyen,
				Doug~Creighton,
        Saeid~Nahavandi
\thanks{
Ngoc Duy Nguyen, Doug Creighton and Saeid Nahavandi are with the Institute for Intelligent Systems Research and Innovation, Deakin University, Waurn Ponds Campus, Geelong, Victoria, Australia (e-mails: duy.nguyen@deakin.edu.au, douglas.creighton@deakin.edu.au and saeid.nahavandi@deakin.edu.au).}
\thanks{
Thanh Thi Nguyen is with the School of Information Technology, Deakin University, Burwood Campus, Melbourne, Victoria, Australia (e-mail: thanh.nguyen@deakin.edu.au).}

}

%
%

\markboth{}%
{Shell \MakeLowercase{\textit{et al.}}: A Visual Communication Map for Multi-Agent\
Deep Reinforcement Learning}
%



\maketitle

\begin{abstract}
Deep reinforcement learning has been applied successfully to solve various real-world problems and the number of its applications in the multi-agent settings has been increasing. Multi-agent learning distinctly poses significant challenges in the effort to allocate a concealed communication medium. Agents receive thorough knowledge from the medium to determine subsequent actions in a distributed nature. Apparently, the goal is to leverage the cooperation of multiple agents to achieve a designated objective efficiently. Recent studies typically combine a specialized neural network with reinforcement learning to enable communication between agents. This approach, however, limits the number of agents or necessitates the homogeneity of the system. In this paper, we have proposed a more scalable approach that not only deals with a great number of agents but also enables collaboration between dissimilar functional agents and compatibly combined with any deep reinforcement learning methods. Specifically, we create a global communication map to represent the status of each agent in the system visually. The visual map and the environmental state are fed to a shared-parameter network to train multiple agents concurrently. Finally, we select the \emph{Asynchronous Advantage Actor-Critic} (A3C) algorithm to demonstrate our proposed scheme, namely \emph{Visual communication map for Multi-agent A3C} (VMA3C). Simulation results show that the use of visual communication map improves the performance of A3C regarding learning speed, reward achievement, and robustness in multi-agent problems.
\end{abstract}


%
\IEEEpeerreviewmaketitle

\section{Introduction}
\label{sec:1}
%
%
%
%

\IEEEPARstart{A}{pplications} of deep reinforcement learning (RL) methods in the multi-agent environments have attracted much attention recently because of their capability of solving various complex problems. Multi-agent learning is to learn the interaction between multiple distributed self-operated agents to balance a shared interest or collectively achieve a designated objective \cite{1, 2}. A vital form of such interaction involves the ability to communicate among agents, which basically ensures agents to function as a team rather than as individuals \cite{3}. Examples of communication activities in multi-agent systems include the scheduling problem in wireless sensor networks \cite{4}, strategic planning in robot soccer \cite{5}, traffic junction problem \cite{6}, elevator control \cite{7}, and multi-agent games \cite{8}. Therefore, learning to communicate in multi-agent systems has drawn a great deal of attention from the research community in recent years \cite{9, 10}.

\begin{figure}[!b]
\centering
\includegraphics[width=0.87\linewidth]{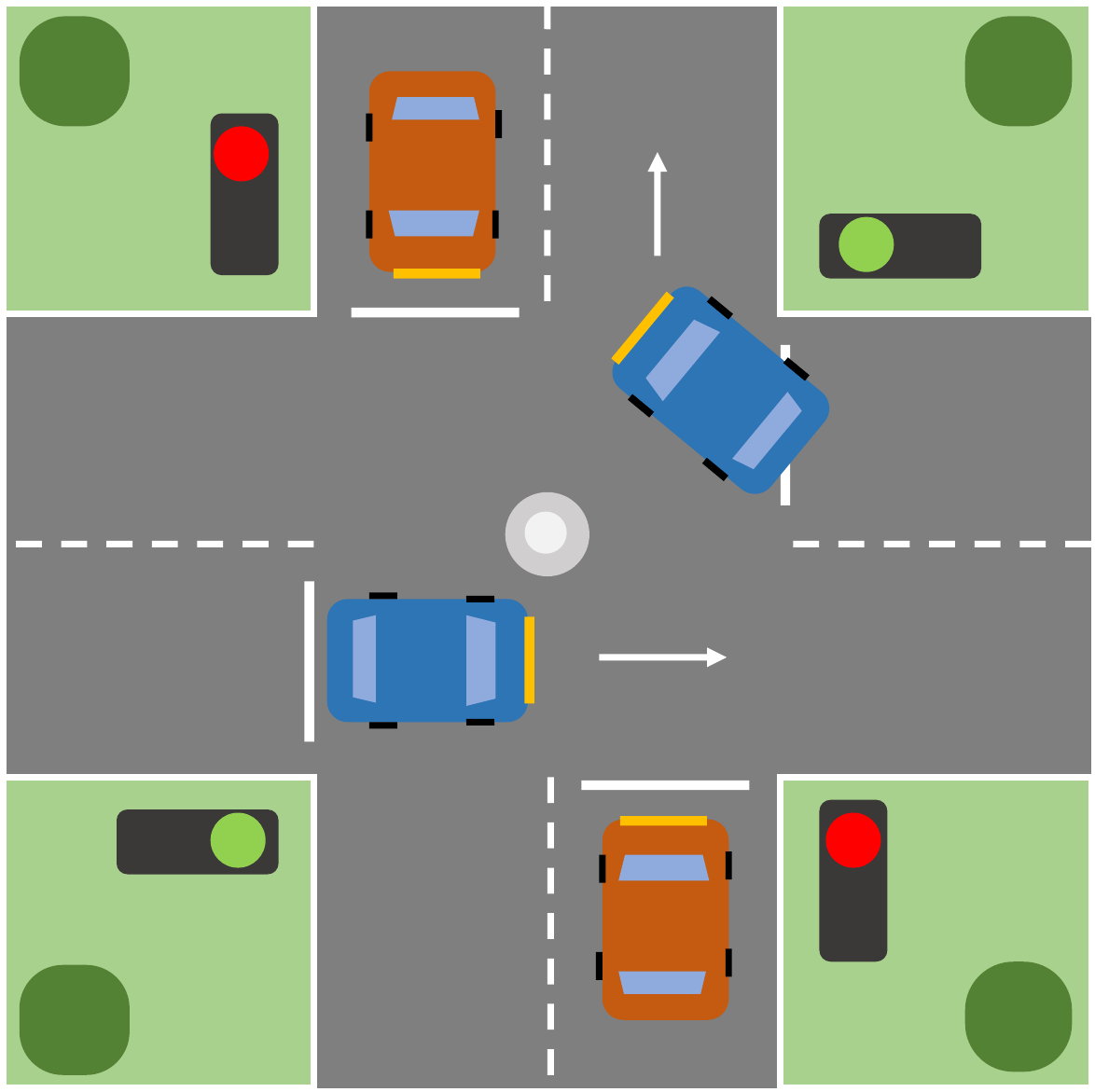}
\caption{A traffic light can be used as a visual indicator to control the level of cooperation between vehicles in a road intersection.}
\label{fig:1} 
\end{figure}

Deep RL \cite{11} is known as a normative approach to deal with the \emph{curse of dimensionality} in domains where the observation space and action space are highly dimensional \cite{12}. It is possible because deep RL approximately estimates the value function by the use of neural networks, which basically include convolutional layers to process raw graphical data directly. In this paper, we leverage the potential of \emph{Convolutional Neural Networks} \cite{13} (ConvNets) to examine the effect of graphical features to the gradual cooperation between interacting agents during training. To make it feasible, we define a set of visual indicators so that each indicator represents the current status of an agent. The visual indicators are globally visible to all agents and plays as a virtual communication medium. For this reason, we call a set of visual indicators a visual communication map.

To clearly present the motivation of the study, we consider the following example. Fig. \ref{fig:1} describes a scenario in a road intersection. Vehicles in the intersection must follow the traffic light to make subsequent decisions and avoid possible collisions. When the light turns red, the vehicles are notified to stop and when the light turns green, the vehicles are allowed to move forward. In this case, the traffic light plays as a virtual communication medium among vehicles. Because the traffic light is visible to every vehicle in the intersection, it can be used as visual indicators to enable cooperation between vehicles. Therefore, the drivers are only required to learn the traffic light rules. The example implies that visual indicators can, somehow, help cooperation in a multi-agent system.

\begin{figure}[!t]
\centering
\includegraphics[width=0.87\linewidth]{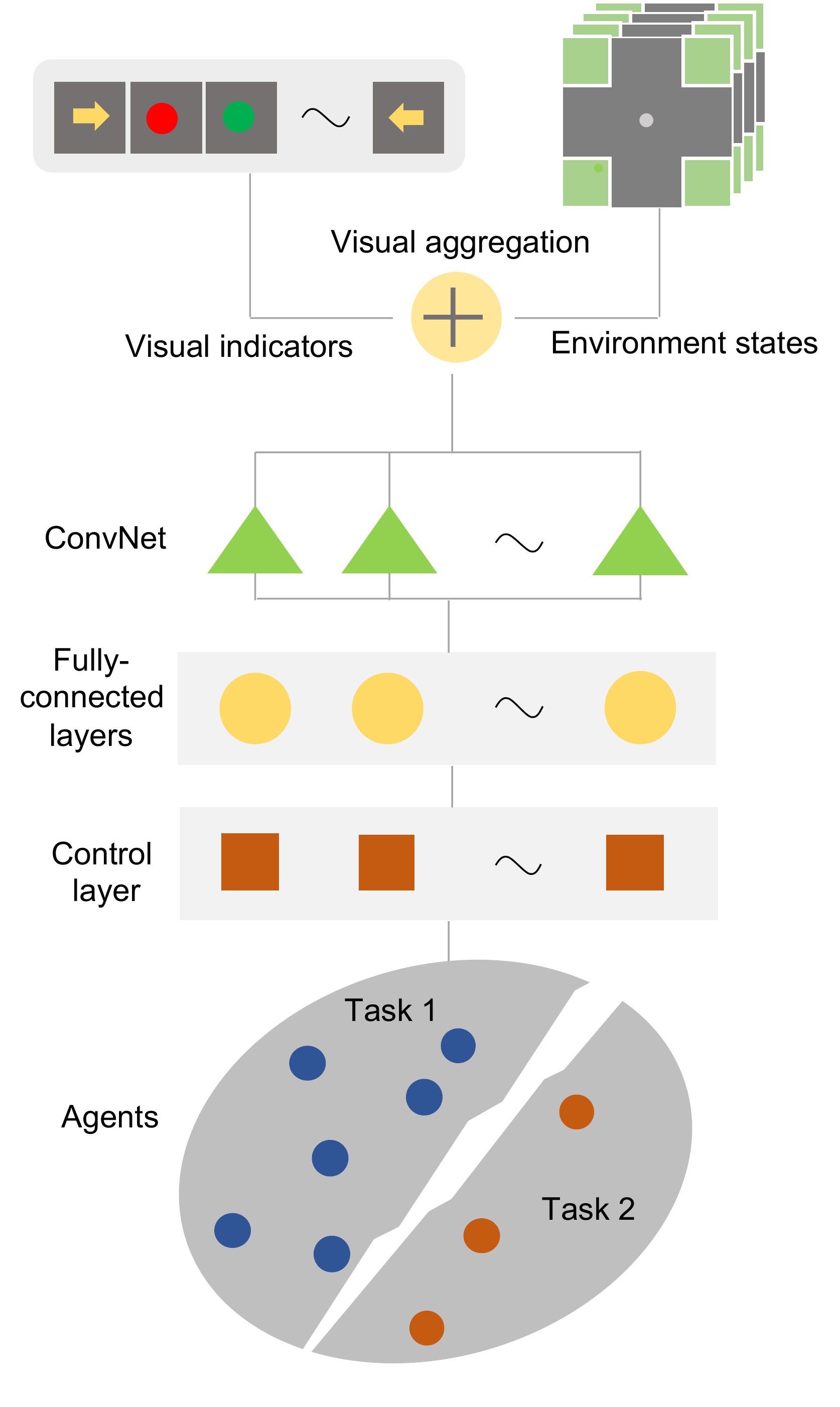}
\caption{Visual indicators are used to enable cooperation between multiple agents.}
\label{fig:2} 
\end{figure}

By analyzing the previous example meticulously, we design a visual communication map for multi-agent environments, as shown in Fig. \ref{fig:2}. Initially, we create a visual map that includes a set of visual indicators. Each indicator represents the current status of an agent. The visual map is combined with the environmental state to feed into a ConvNet. The output of the ConvNet is connected with a fully-connected network followed by a control layer where subsequent actions are predicted. Finally, each agent uses the augmented knowledge of other agents via the visual communication map to learn the optimal distributed policy on its own and, at the same time, maintain a certain level of cooperation with other agents.

Deep RL brings in a great deal of success in many complex real-world problems such as Atari games \cite{12, 14, 14b, 14c}, the game of Go \cite{15}, robotics control tasks \cite{16, 17, 18, 18b}, autonomous driving \cite{19}, and surgical robotics \cite{19b, 19c}. As opposed to previous studies, we examine deep RL in multi-agent domains. Multi-agent learning is more sophisticated than the single-agent counterpart as the action space exponentially increases with the number of agents. Furthermore, we face the challenge of \emph{moving target problem} \cite{21, 22}, which is a primary difficulty in finding the optimal policy of every agent. In other words, the rewarding signal of an agent depends on the decisions of other agents and causes the seeking policies to become \emph{non-stationary} \cite{22, 23}. The use of \emph{experience replay} memory even makes the situation worse as a large portion of transitions in the memory could become deprecated \cite{21, 24, 25}. For this reason, we choose the A3C \cite{26} method, which is a policy-based algorithm, to demonstrate our proposed scheme. There are different variants of A3C such as \cite{27, 28, 29}, but we use the A3C version based on GPU \cite{27} due to various beneficial factors: high performance with multiple concurrent actor-learners, efficiency within a short period of training, robustness with a wide range of learning rate, and easy to implement due to many open-source codes available in Python.

\begin{figure}[h]
\centering
\includegraphics[width=0.8\linewidth]{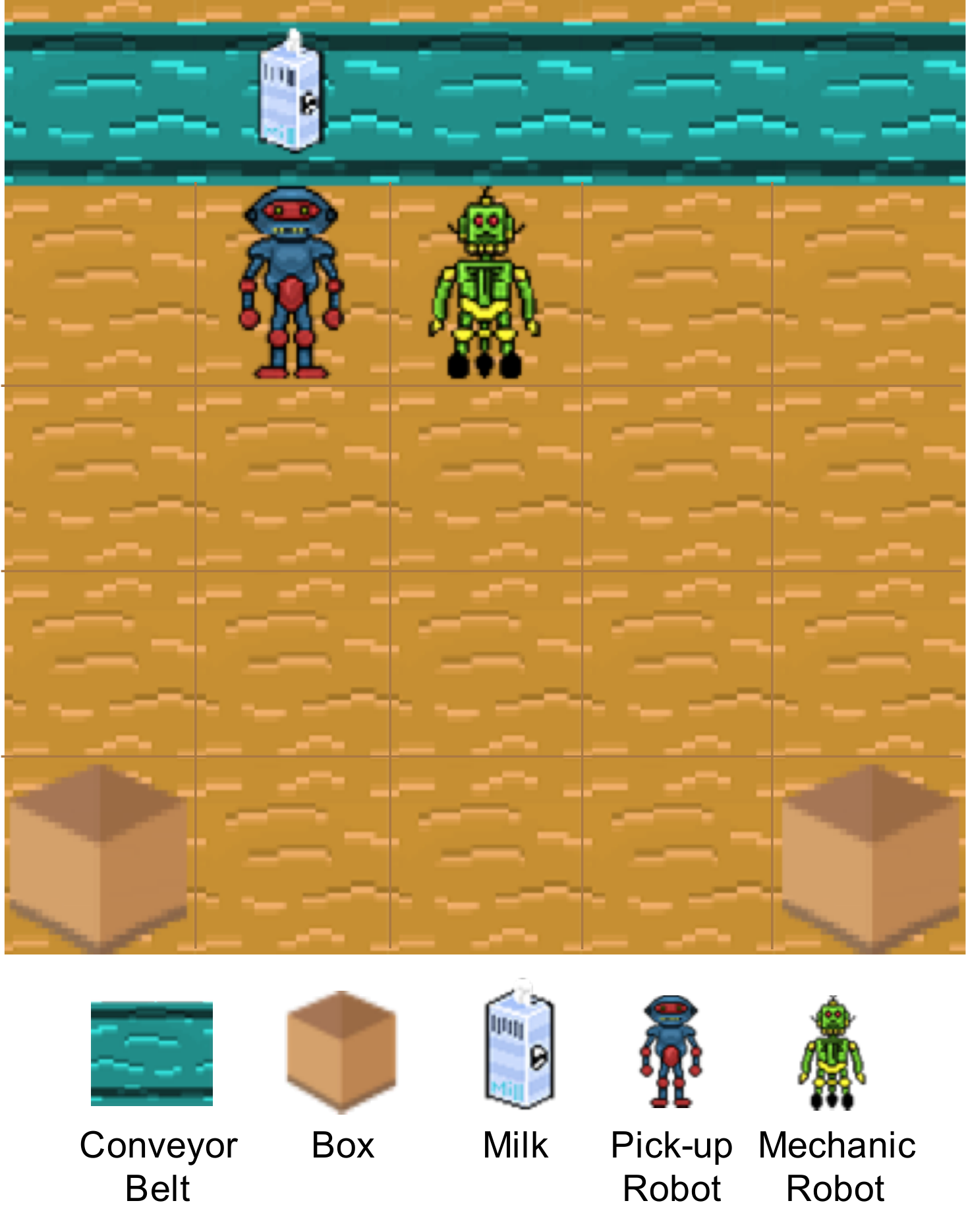}
\caption{A gameplay interface of Milk Factory.}
\label{fig:3} 
\end{figure}

Last but not least, we develop a game named \emph{Milk Factory} that is used to benchmark our proposed scheme. Milk Factory simulates a task in a warehouse where robots are used to automate the production chain. Fig. \ref{fig:3} describes the gameplay interface of Milk Factory. There are two kinds of robots in the factory: pick-up robots and mechanic robots. At first, a pick-up robot waits for the milk bottle that is on the running conveyor belt and picks it when the bottle is in front of the robot. The robot can only pick one milk bottle at a time. It then brings the bottle to the box that is placed far away from the conveyor belt. After placing the milk bottle in the box, the robot is free and can go back to the conveyor belt to wait for another milk bottle. However, the pick-up robot may stop to work (e.g. due to battery depletion) during the operation period. In this case, the mechanic robot goes to the pick-up robot's position and fixes the error. After fixing, the pick-up robot can continue to operate. A pick-up robot is rewarded a score of 10 whenever it picks a milk bottle or puts the bottle in the box. The mechanic robot is also given a score of 10 whenever it fixes the pick-up robot successfully. Therefore, the game requires the cooperation of two robots to maximize reward achievement.

Finally, the paper contributes to the following highlights:

\begin{itemize}

\item The study conducts an interesting investigation of the relationship between visual indicators and interacting agents. Experimental results show that visual indicators play as a communication bridge to enable the cooperation between multiple agents. Therefore, the study provides a practical method for the development of multi-agent systems.

\item We develop a multi-agent environment named Milk Factory. Milk Factory is developed in Python, which follows the standard interface of \emph{Arcade Learning Environment} (ALE) \cite{30, 31} and \emph{OpenAI Gym} \cite{32}. Therefore, previous deep RL methods, which work with ALE and OpenAI Gym, adapt well with Milk Factory without significant modifications. Furthermore, Milk Factory is a configurable game that provides unlimited potentials to suit with any research purposes.

\item A3C is not a well-known method for multi-agent learning \cite{8}. In this study, however, we show that the combination of visual communication map and A3C can work properly with multi-agent problems.

\item Although we use A3C to demonstrate our proposed scheme, it can be used with any deep RL algorithms. Furthermore, the method is scalable in terms of the number of agents and compatible with heterogeneous environments where agents can function different tasks.

\end{itemize}

The paper is organized as follows. Section \ref{sec:2} examines recent studies in multi-agent domains. Section \ref{sec:3} presents the concept of a visual communication map and the proposed scheme VMA3C. The performance evaluation is conducted based on Milk Factory in Section \ref{sec:4}. Finally, to conclude the paper, we present potential extension directions and the future work in Section \ref{sec:5}.

\section{Related Work}
\label{sec:2}

Since the first development of deep RL, namely \emph{Deep Q-Network} (DQN) \cite{12}, there have been numerous variants of DQN such as \emph{Double Q-Learning} \cite{25}, \emph{Dueling Network Architecture} \cite{33}, \emph{Recurrent Deep Q-Learning} \cite{34}, \emph{Deep Attention Recurrent Q-Network} \cite{35}, and \emph{Prioritized Experience Replay} \cite{24}. However, these approaches concentrate on approximating the value function, which requires a large amount of memory to store historical transitions. Furthermore, the experience replay is known to amplify the non-stationary problem in multi-agent systems \cite{21}. Therefore, a policy-based approach such as A3C \cite{26} and its variants, e.g. \emph{UNsupervised REinforcement and Auxiliary Learning} (UNREAL) \cite{36}, are developed to hasten the learning speed by allowing multiple actors-learners to be trained at the same time. These methods are more efficient than the value-based methods as they do not require an experience replay. Simulation results show that the policy-based methods outperform the value-based ones concerning learning speed and total reward achievement in the Atari domain \cite{26}. Besides, \emph{Deep Deterministic Policy Gradient} (DDPG) \cite{37} and \emph{Trust Region Policy Optimization} (TRPO) \cite{38} are proposed to deal with continuous action domains. In the first case, DDPG combines DQN with the actor-critic architecture to find the optimal policy. Whereas, TRPO is a policy gradient method that allows controlling the policy improvement in every training step. In this paper, we select A3C as the baseline method to demonstrate our proposed scheme because Milk Factory has a discrete action space.
                                                                                                                                                                                                                                                                                                                                                                                                         
There are notable methods that use communication to control multiple agents such as Foerster \emph{et al.} \cite{3} formulate two approaches \emph{Reinforced Inter-Agent Learning} (RIAL) and \emph{Differentiable Inter-Agent Learning} (DIAL). In the first scheme, RIAL combines DQN with a recurrent network to independently train every agent through a shared-parameter network.  Whereas, DIAL additionally sends messages across agents during the centralized learning. These real-valued messages are mapped to a limited number of communication actions. In this paper, we also adopt the centralized learning by training multiple agents at the same time but operating in a distributed nature when deployed. It is feasible because the visual indicators can be used as a virtual communication medium. Agents are trained to learn the cooperation through the visual indicators. Because the channel is virtual, there is no cost for exchanging messages between agents and hence energy efficiency.

Sukhbaatar \emph{et al.} \cite{6} propose a communication model that allows multiple agents to communicate before deciding subsequent actions. The authors create a deep neural network that has an open access to a shared communication channel. Agents receive aggregated transmissions from other agents via a continuous vector embedded in the channel. Because the communication channel carries continuous data, it is possible to combine the approach with a standard single-agent RL method. However, the scheme is not scalable due to the complication of the neural network structure. In this study, we use the visual communication map, which is basically separated from the network configuration. Therefore, our proposed method is more robust and scalable.

Finally, Gupta \emph{et al.} \cite{8} examine a set of cooperative control tasks and compare the performance of three different deep RL approaches policy gradient, temporal-difference error, and actor-critic method. The authors combine various disciplines to efficiently train multiple agents to learn a wide range of control tasks from discrete to continuous action domains. Initially, the agents are trained to learn a simple task and move to the harder ones afterward. The knowledge of the previous task is accumulated over time rather than acquiring new knowledge from scratch. This approach is called \emph{curriculum learning} \cite{39, 40}. The authors also show that the use of a shared-parameter network is scalable in terms of the number of agents. However, the paper examines only homogeneous domains where agents function similar jobs. In our study, we extend the investigation to heterogeneous systems where agents can be in different task domains. Last but not least, we though evaluate our proposed framework in Milk Factory, a fully observation problem, the presented approach can be applied to partially observation domains such as \cite{41} similarly.

\section{Proposed Schemes}
\label{sec:3}

In this section, we have first presented preliminary terminologies of RL and the A3C method. We then introduce the use of a visual communication map with A3C as well as the network architecture for multi-agent problems. Finally, we propose two implementation approaches for the visual communication map.

\subsection{Preliminary}
\label{sec:3.1}

\subsubsection{RL}

\begin{figure}[!t]
\centering
\includegraphics[width=0.9\linewidth]{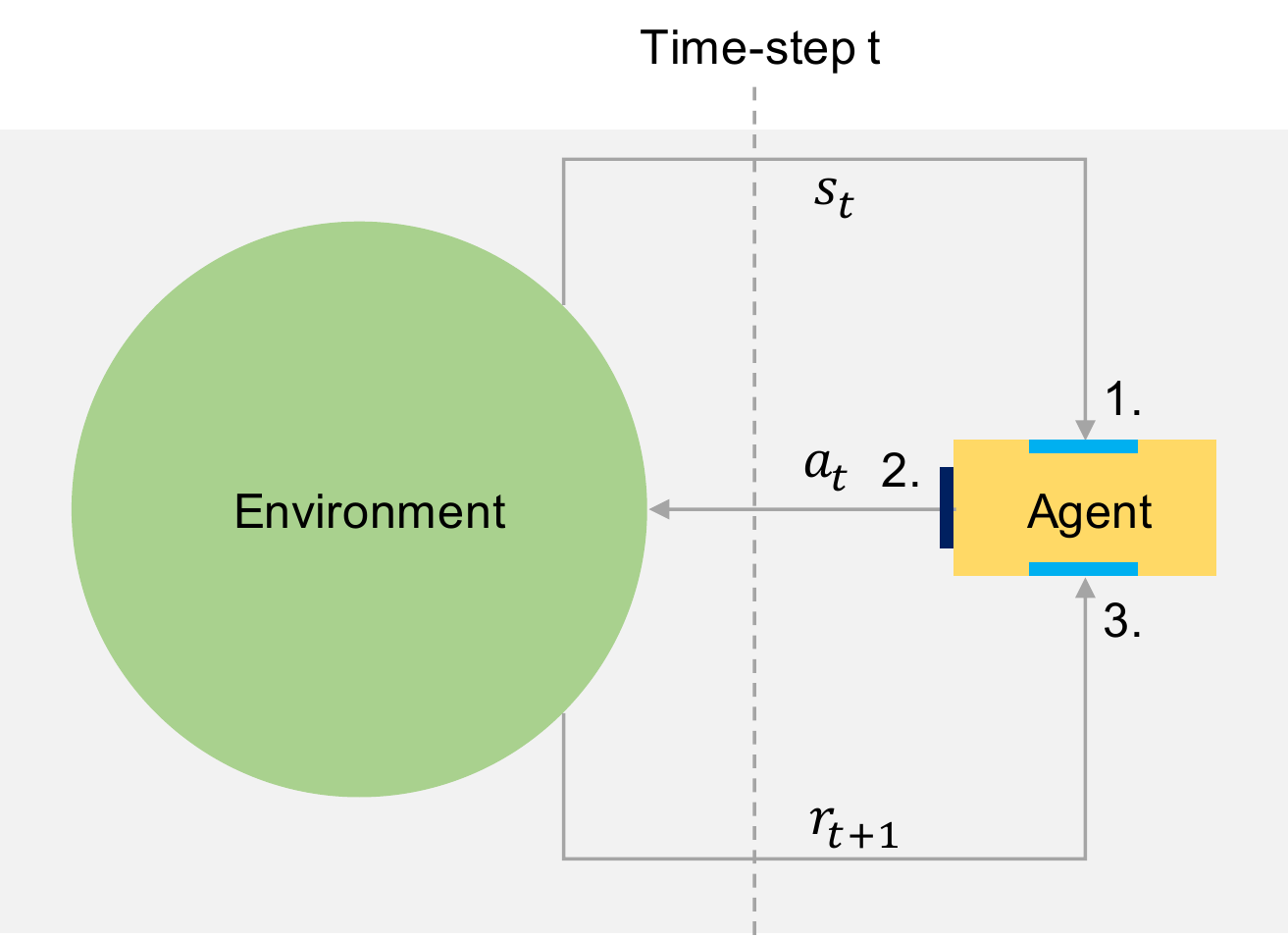}
\caption{The relationship between an agent and the environment in RL.}
\label{fig:4} 
\end{figure}

RL involves the interaction between an agent and the environment, as shown in Fig. \ref{fig:4}. At time-step $t$, the agent perceives an environmental state $s_t$. It reacts with an action $a_t$ and the environment returns a reward $r_t$. From time to time, the agent obtains a set of transitions: $S^{T} = \{s_0, a_0, r_1, s_1, a_1, r_2, ..., s_T, a_T, r_{T+1} \}$, where $T$ is the terminal state of the episode. The initial state $s_0$ is generated randomly from the observation space $S$ and the terminal reward $r_{T+1}$ is assigned to 0.

On the agent's side, a policy $\pi$ is defined as a mapping function $f$ from the observation space $S$ to the action space $A$, \emph{i.e.}, $\pi = f: S \longrightarrow A$. In stochastic environments, it is convenient to define the policy $\pi$ as a conditional probability distribution of action $a_t$, given the state $s_t$, as follows:

\begin{equation*}
\pi(a_t|s_t) = \left\lbrace\ p(a_{t}|s_t)\ \bigg\vert\ \forall a_{t} \in A, \forall s_t \in S\right\rbrace.
\end{equation*}

In this definition, the next state $s_{t+1}$ is assumed to depend only on the previous state $s_t$ and the corresponding action $a_t$. A problem that satisfies this condition is called a \emph{Markovian decision process} (MDP). Milk Factory is an MDP. In real-world problems, the agent only senses a limited view of the surrounding environment. Therefore, it is more practical to formulate the problem as a \emph{partially observable MDP} (POMDP) \cite{42}. In this paper, for concise, we only consider the MDP, but the same procedure can be inferred for the POMDP.

On the other hand, we define the discounted return at time-step $t$ as $R_t = r_{t+1} + \gamma r_{t+2} + \gamma^2 r_{t+3} + ... $, where $\gamma$ is a discounted factor so that $\gamma \in [0, 1]$. The goal of RL is to maximize the discounted return $R_t$.

Finally, to evaluate the \emph{value} of a state, we define the value function $V$ of a state $s_t$ under a policy $\pi$ as the expected value of the discounted return $R_t$, \emph{i.e.},

\begin{equation}
V(s_t)|_{\pi}=\mathbf{E}\left\lbrace R_t\middle\vert \pi \right\rbrace = \mathbf{E}\left\lbrace \sum_{i=0}^{T-t-1}\gamma^{i}r_{t+i+1}\middle\vert \pi\right\rbrace.
\label{eq:1}
\end{equation}

\subsubsection{A3C}

A3C is a policy-based method that uses the actor-critic architecture \cite{43} and the advantage function \cite{44} to estimate the optimal policy $\pi*$. Furthermore, it enables concurrent learning by creating multiple actors-learners to update the parameters of the neural network asynchronously. In particular, as shown in Fig. \ref{fig:5}, the A3C's network structure consists of two essential sub-networks: one actor network $N^a(\theta)$ and one critic network $N^c(\theta')$, where $\theta$ and $\theta'$ denote $N^a$'s weight parameters and $N^c$'s weight parameters respectively. The actor network estimates the policy $\pi(a_t|s_t, \theta)$ of the agent while the critic network evaluates the value function $V(s_t, \theta')$ (\ref{eq:1}). Therefore, the A3C method optimizes two loss functions: the actor loss function $L^a(\theta)$ and the critic loss function $L^c(\theta')$: 

\begin{figure}[!h]
\centering
\includegraphics[width=0.9\linewidth]{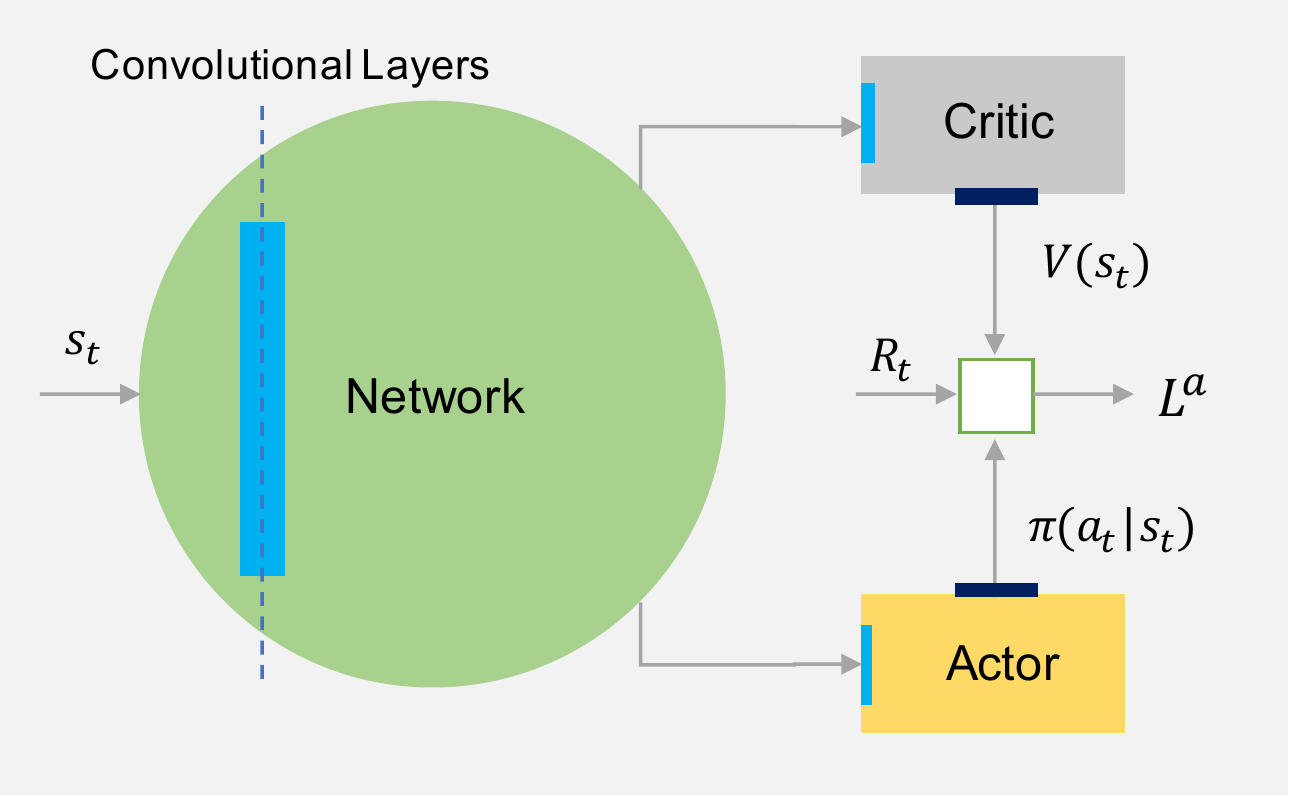}
\caption{Actor-critic architecture in A3C.}
\label{fig:5} 
\end{figure}

\begin{equation}
L^a(\theta) \simeq \log(\pi(a_t|s_t,\theta)) \left(A_t(\theta') - V(s_t, \theta')\right) 
\label{eq:2}
\end{equation}

\noindent
and

\begin{equation}
L^c(\theta') \simeq \frac{1}{2}\left( R_t(s_t, \theta') - V(s_t, \theta')\right)^2,
\label{eq:3}
\end{equation}

\noindent
where $A_t$ is calculated by the following equation:

\begin{equation}
A_t(\theta') = \sum_{k=0}^{T_{max}-t-1}\gamma^k r_{t+k+1} + \gamma^{T_{max}-t} V(s_{T_{max}}, \theta'),
\label{eq:4}
\end{equation}

\noindent
where $T_{max}$ denotes the maximum number of observation steps. If $T_{max} = T$, we have $V(s_{T_{max}}, \theta') = V(s_{T}) = 0$. Finally, to enable the agent's exploration during the training process, the entropy regularization factor $H(s_t,\theta)$ is added to the total loss.

Finally, the total loss is the summation of the actor loss, the critic loss, and the entropy regularization:

\begin{equation*}
L^{total} = L^a + L^c + \beta H,
\end{equation*}

\noindent
where $\beta$ denotes the control parameter so that $\beta \in [0,1]$.

\subsection{VMA3C}

\subsubsection{Multi-agent A3C}

\begin{figure}[!h]
\centering
\includegraphics[width=0.9\linewidth]{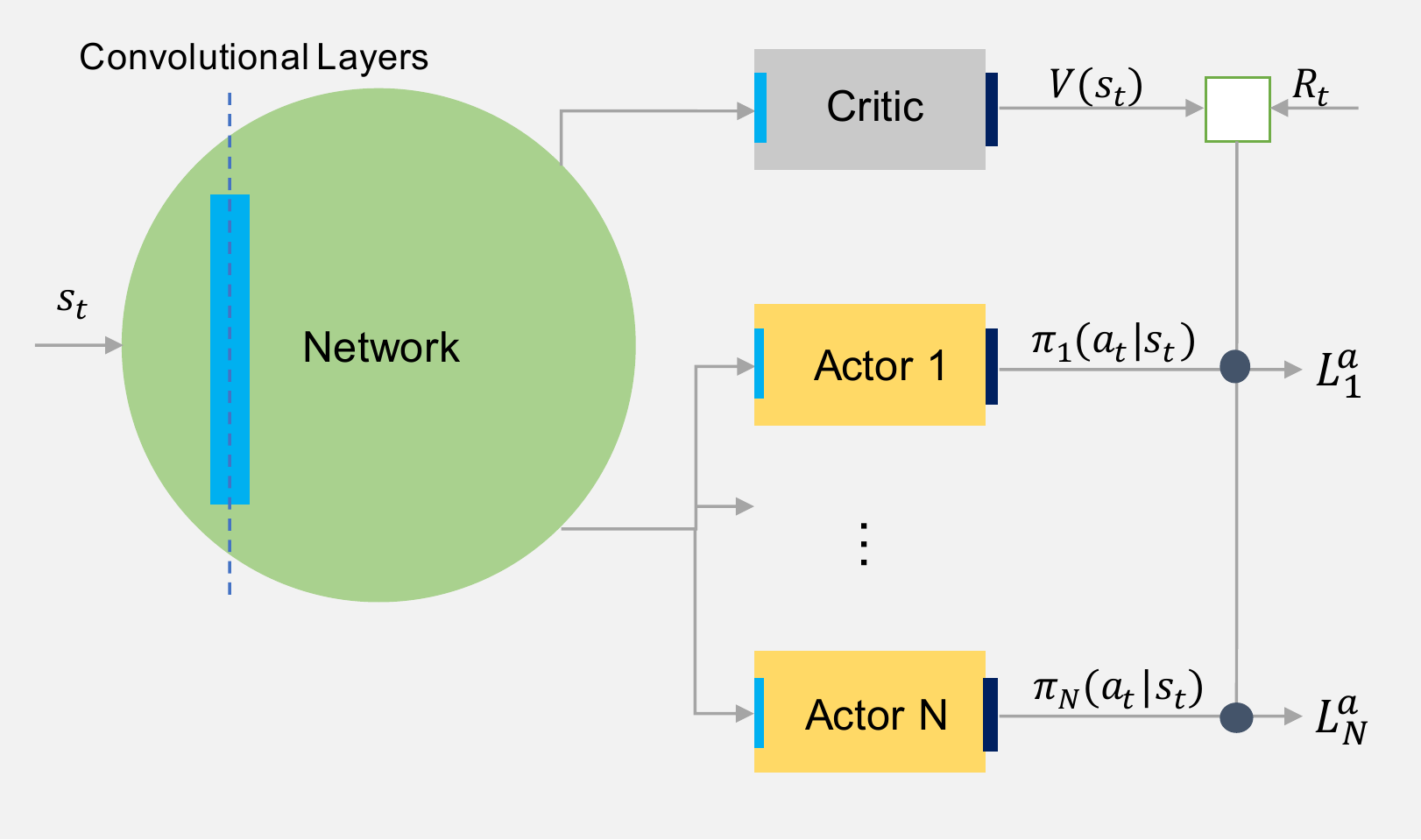}
\caption{Multiple-actor-single-critic architecture in A3C.}
\label{fig:6} 
\end{figure}

In this section, we extend the use of A3C to multi-agent learning, as shown in Fig. \ref{fig:6}. The procedure can be applied to any standard deep RL methods. It is even more straightforward to derive from a value-based method such as DQN because its network has a single output while the A3C's network includes two outputs (actor-critic), as explained in the previous subsection.

We consider an MDP (or a POMDP) that consists of $N$ agents $\{1, 2, ..., N\}$. Because A3C has two factors in the network configuration (actor and critic), we create $N$ actor networks $N^a_i(\theta_i)$ for $N$ agents ($i=1..N$) while using a single critic network $N^c(\theta')$. To efficiently train multiple agents at the same time, we use a shared-parameter network for $N$ actors, \emph{i.e.}, $\theta_1 = \theta_2 = ... = \theta_N = \theta$. Let $\pi_i(a_{i,t}|s_t)$ be a policy of the agent $i$, where $a_{i,t} \in A_i$, $s_t \in S$, and $A_i$ is the action space of the agent $i$. We consider that the intermediate reward $r_t$, given $s_t$ and $a_{i,t} (i=1..N)$, is the summation of individual reward, \emph{i.e.}, $r_t = \sum_{i=1}^N r_{i,t}$. Because the state space $S$ is shared among all agents and because we use a single critic network, the critic loss function for a multi-agent scenario is the same as in equation (\ref{eq:3}), \emph{i.e.}, $L^c(\theta') \simeq \frac{1}{2}\left( R_t(s_t, \theta') - V(s_t, \theta')\right)^2$. The actor loss  $L^a_i(\theta)$ of the agent $i$ is also calculated as in equation (\ref{eq:2}) as below:

\begin{equation*}
L^a_i(\theta) \simeq \log(\pi_i(a_{i,t}|s_t,\theta)) \left(A_t(\theta') - V(s_t, \theta')\right). 
\end{equation*}

\noindent
However, the entropy regularization $H$ in this case is the summation of $N$ individual factors, \emph{i.e.}, $H = \sum_{i=1}^N H_i(s_t,\theta)$. 

Finally, we have the total loss for multi-agent problems using A3C as follows: 

\begin{equation}
L^{total} =  \sum_{i=1}^N L_i^a + \alpha L^c + \beta H,
\label{eq:5}
\end{equation}

\noindent
where $\alpha$ ($\alpha \geq 1$) is added to balance the importance of the critic among multiple actors. 

\subsubsection{Visual communication map}

For each agent, we define a set of status $S^{stat}_i = \{c_{i1}, c_{i2}, ..., c_{ij}\}$, where $i=1..N$, $j \geq 0$, and $c_{il} \neq c_{ip} (l \neq p)$. We also define a mapping function $F_i$ that maps $S^{stat}_i$ to a set of graphical representations $G = \{g_1, g_2, ..., g_m\}$ so that 

\begin{itemize}
\item $\forall x \in s_t \Rightarrow x \not \in g_i$,
\item $\forall x \in g_i \Rightarrow x\not\in g_j (i \neq j$ and $i,j \in \{1,...,m\})$,
\end{itemize}

\noindent
where $m = \sum_{i=1}^N ||S^{stat}_i|| $. In other words, we have $F_i: S^{stat}_i \rightarrow G$ $(i=1..N)$.

Let $C_i^t$ be the status of the agent $i$ at time-step $t$. We have $C_i^t \in S^{stat}_i$. A visual communication map at time-step $t$ is defined as a set of graphical representations of $N$ agents' status at time-step $t$: $M_t = \{F_1(C_1^t), F_2(C_2^t), ..., F_N(C_N^t)\}$.

Finally, we define an operator $\oplus$ that combines the environmental state $s_t$ with $M_t$. We assume that $S_t = s_t \oplus M_t$. Then, $S_t$ is used to feed to the neural network (instead of $s_t$). To maximize the learning efficiency, we define the operator $\oplus$ so as to satisfy the following conditions:

\begin{frm-thm}
The environmental state $s_t$ is a strict subset of $S_t$, i.e., $s_t \subset S_t$.
\end{frm-thm}

\begin{frm-thm}
Every element of the visual communication map $M_t$ is a strict subset of $S_t$, i.e., $F_i(C_i^t) \subset S_t (i=1..N)$.
\end{frm-thm}

\begin{frm-thm}
Given $E$=$\{s_t, F_1(C_1^t), F_2(C_2^t), ..., F_N(C_N^t)\}$ and an arbitrary $x$ so that $x \in S_t$, we have that $x$ exclusively belongs to one element of $E$.
\end{frm-thm}

Premises 1 and 2 make sure that $S_t$ maintains the integrity information of the environmental state and the current status of all agents in the system. Premise 3 is defined to maximize learning efficiency. We can infer the following rules based on these conditions:

\begin{frm-lmm}
Let $s_t \neq \emptyset$ be the environmental state at time-step $t$ and $G_i^t$ be the graphical representation of the status of agent $i$ at time-step $t$, i.e., $G_i^t = F_i(C_i^t) (i=1..N)$ so that $\exists k \in \{1, ..., N\}$ and $G_k^t \neq \emptyset$. The following operator $S_t = s_t \oplus M_t$ $=$ $s_t \cup \left( \cup_{i=1}^N G_i^t \right)$ satisfies the three premises.
\end{frm-lmm}

\noindent
\emph{Proof}: Because $S_t = s_t \cup \left( \cup_i^N G_i^t \right)$, we have $s_t \subseteq S_t$ and $G_i^t \subseteq S_t (i=1..N)$. However, the \emph{equal} sign does not occur because $s_t \neq \emptyset$ and $\exists k \in \{1, ..., N\}$ so that $G_k^t \neq \emptyset$ and $s_t \neq G_i^t (i=1..N)$. This implies that the operator satisfies premises 1 and 2. 

Secondly, we get an element $x \in S_t$. We assume that $x$ belongs to at least two elements of $E$. There are two possible cases: 

\begin{itemize}
\item If $x \in s_t$, we have $\exists k \in \{1, ..., N\}$ so that $x \in G_k^t$ and $G_k^t \neq \emptyset$. This contradicts with the definition of the visual communication map, \emph{i.e.}, $x \not \in G_k^t (\forall x \in s_t)$.

\item If $x \not\in s_t$, we have $\exists i,j \in \{1, ..., N\}$ so that $i \neq j, x \in G_i^t$ and $x \in G_j^t$, which also contradicts with the definition of the visual communication map.
\end{itemize}

The lemma is completely proved. $\hfill \blacksquare$

\begin{frm-lmm}
Let $s_t \neq \emptyset$ be the environmental state at time-step $t$, $G_i^t$ be the graphical representation of the status of agent $i$ at time-step $t$, and $\Phi$ be an arbitrary 1-to-1 mapping function from set to set. We assume that $G_i^t \neq \emptyset (i=1..N)$. The following operator $S_t = s_t \oplus M_t$ $=$ $\Phi(s_t) \cup \left( \cup_{i=1}^N \Phi(G_i) \right)$ satisfies the three premises.
\end{frm-lmm}

\noindent
\emph{Proof}:  Because $\Phi : A \rightarrow B$ is a 1-to-1 mapping function from set $A$ to set $B$, \emph{i.e.}, $\forall x \in A, \exists! y \in B \Rightarrow y = \Phi(x)$ and vice versa. Therefore, the problem can be converted to Lemma 1, which is completely proved. $\hfill \blacksquare$

Lemma 1 and Lemma 2 are important as they can be used to suggest different implementation approaches for the visual communication map. In this paper, we suggest the following approaches:

\begin{figure}[!h]
\centering
\includegraphics[width=0.9\linewidth]{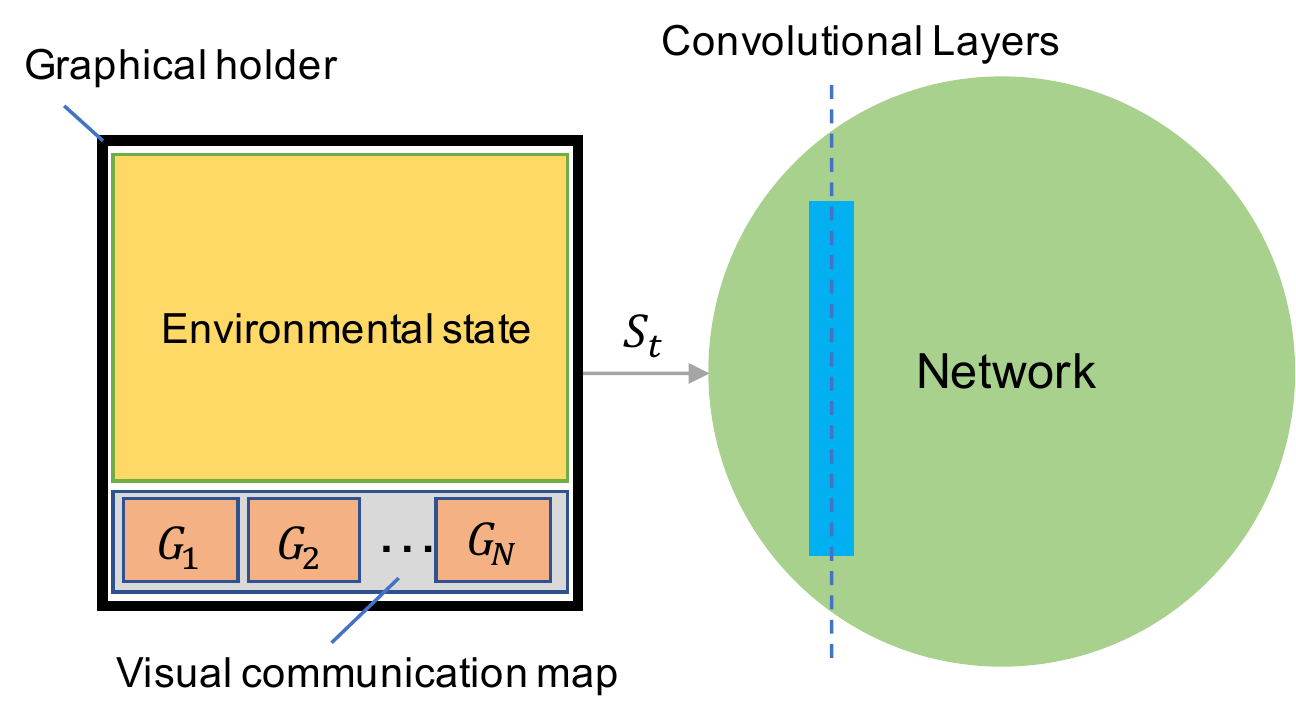}
\caption{Visual aggregation by creating a holder.}
\label{fig:7} 
\end{figure}

\begin{figure}[!t]
\centering
\includegraphics[width=0.9\linewidth]{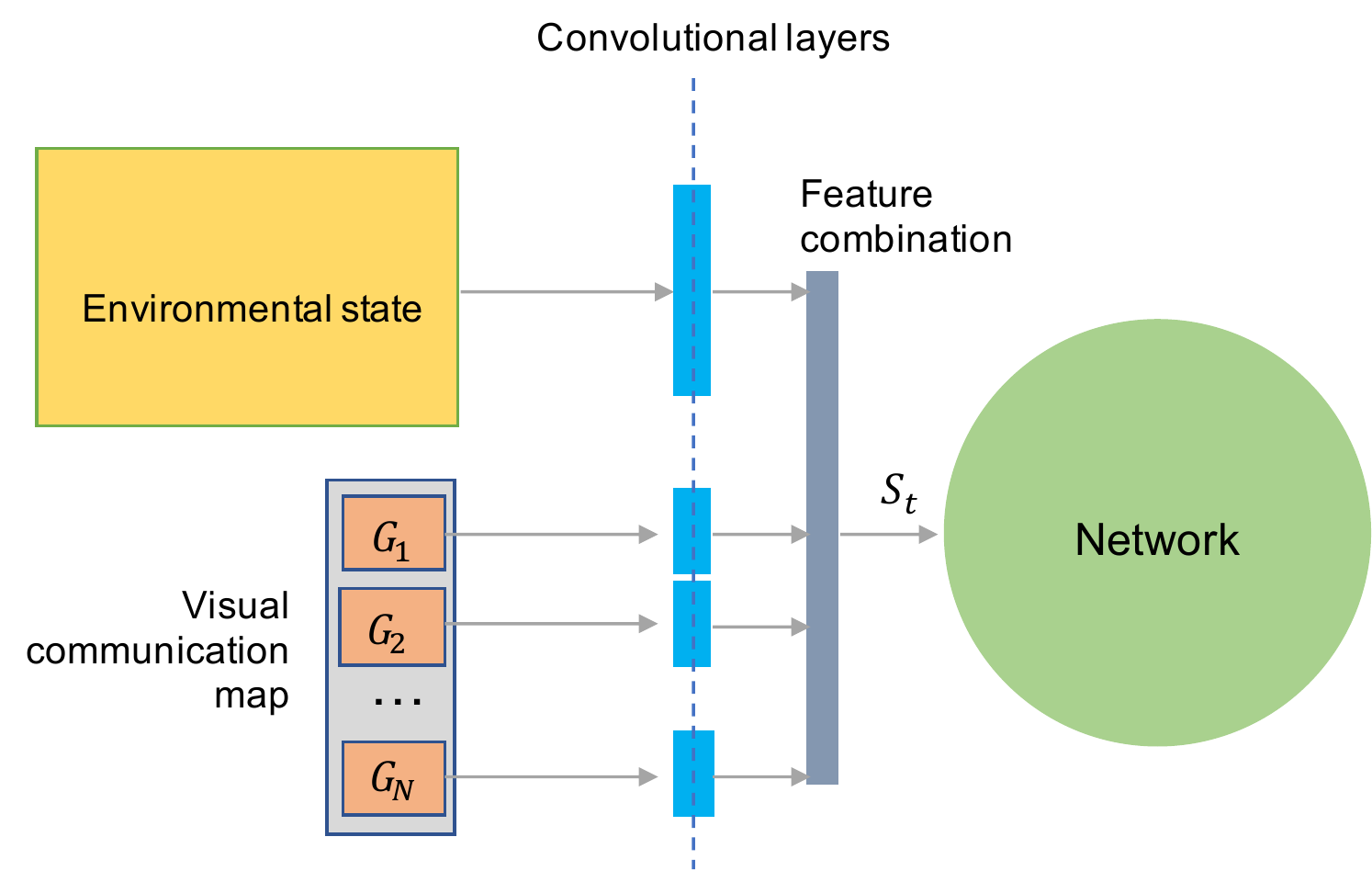}
\caption{Concatenating output features of $N+1$ ConvNets.}
\label{fig:7b} 
\end{figure}

\begin{enumerate}
\item We create a graphical holder (a mask) that embeds both the environmental state and the visual communication map, as shown in Fig. \ref{fig:7}. The communication map is a graphical representation that contains the current status information of $N$ agents. The holder is used as $S_t$ and is fed to the neural network. This approach is inferred from Lemma 1.

\item We create $N+1$ ConvNets: one for the environmental state and the other $N$ ConvNets for $N$ agents, as shown in Fig. \ref{fig:7b}. The output features of $N+1$ ConvNets are concatenated to form $S_t$. This approach is inferred from Lemma 2.
\end{enumerate}

In this paper, we use the first method to benchmark our proposed scheme because it is more efficient (using a single shared-parameter ConvNet). The detailed algorithm is described in \textbf{Algorithm} \ref{alg}.

\begin{algorithm}[!h]
\caption{Visual communication Map for Multi-agent A3C (VMA3C)}\label{alg}
\small
\begin{algorithmic}[1]
\State $N \gets 0$ \Comment{Global shared counter}
\Procedure{VMA3C}{} \Comment{A learner procedure}
\State $t \gets 0$
\Repeat 

\State $s \gets s_t \oplus M_t$
\State $c \gets 0$ 
\State Reset experience memory $E \gets \{\}$

\Repeat
\State $a \gets \{a_{1,t}, a_{2,t}, ..., a_{N,t}\}$ \Comment{Using $\pi_1, \pi_2, ..., \pi_N$}
\State $r \gets r_{t+1} = \sum_i^N r_{i,t+1}$ 
\State $s' \gets s_{t+1}$
\State Save transition $(s, a, r, s')$ to $E$ 
\State $s \gets s'$
\State $t \gets t+1, N \gets N+1, c \gets c+1$

\Until{$s = s_T \text{ or } c = T_\text{max}$ }

\State Retrieve all $(s_j,a_j,r_j,s'_j)$ from $E$
\State Calculate $A_t$ based on (\ref{eq:4})
\State Calculate gradients $\frac{\partial L^{total}}{\partial \theta}$, $\frac{\partial L^{total}}{\partial \theta'}$ based on (\ref{eq:5})
\State Perform asynchronous update of $\theta$ and $\theta'$

\If{$s=s_T$}
\State $t \gets 0$
\EndIf

\Until{$N > N_\text{end}$}
\EndProcedure
\end{algorithmic}
\end{algorithm}

In a real-world application, it is not scalable if the number of visual representations increases with the number of agents. Therefore, it is more efficient to use the following \emph{condensed} rule to reduce the number of visual representations. 

\noindent
\textbf{Condensed rule:} Let $c_{ij}$ be the status $j$-th of an agent $i$. If there exists $K = \{k_1, k_2, ..., k_{N-1}\}$ so that the conditional probability $P(c_{ij} | c_{m k_p}) > 0$ $(\forall m \in \{1, 2, ...,N\} - \{i\}, k_p \in K)$ then $F_i(c_{ij}) = \emptyset$.

For example, in the road intersection problem, we only need three visual indicators (red, green, orange) to control all vehicles. However, this rule requires extra information about the problem to estimate the conditional probability $P(c_{ij} | c_{m k_p})$. 

\section{Performance Evaluation}
\label{sec:4}

\subsection{Parameter settings}
\label{sec:4.0}

In this section, we describe the parameter settings that are used in Milk Factory and the proposed scheme (VMA3C) for performance evaluation. The algorithm is run in a computer with an octa-core processor and a GTX 1080Ti graphics card. In Milk Factory, an episode terminates when the number of steps reaches 200. The pick-up robot is rewarded a score of 10 whenever it successfully picks a milk bottle or puts it into the box. Whereas, the mechanic robot is rewarded a score of 10 if it successfully fixes the pick-up robot. The error rate of the pick-up robot is represented as a random variable \emph{ER}, \emph{i.e.}, the probability of an error per action of the pick-up robot equals to $1\%$. Moreover, the pick-up robot has 5 decision actions: moving up, moving down, moving left, moving right, and picking/dropping a milk bottle. The mechanic robot also has 5 decision actions: moving up, moving down, moving left, moving right, and fixing the pick-up robot. A history of 4 game frames is used as a single state to feed into the neural network. 

Finally, based on the \emph{condensed rule}, we only map the status of the pick-up robot to visual representations because the state of the mechanic robot depends on the pick-up robot. The pick-up robot has two states: busy state (bringing the milk bottle) and failed state (because of errors). A milk bottle represents the visual indicator of a busy state and a question mark represents the visual indicator of a failed state.

We compare the performance of the VMA3C method with the A3C method. We keep the parameter settings of A3C same as the previous study \cite{26} except the following changes. The learning rate starts from 0.001 and anneals to 0 during the course of training. The discounted factor $\gamma$ equals to 0.99. The maximum observation step $T_{max}$ equals to 5. The encouraging factor $\beta$ equals to 0.01. Moreover, we perform the gradient clipping by using the following formula \cite{45}:

\begin{equation*}
\delta_i^G = \delta_i^G * \frac{\varpi}{\max(\omega, \varpi)},
\end{equation*}

\noindent
where $\varpi=40$, $\forall i:\delta_i^G \in \Delta_G$, and $\forall j: \delta_{i,j}^G \in \delta_i^G$ we have:

\begin{equation*}
\omega = \sqrt{\sum_{i,j} |\delta_{i,j}^G|^2}.
\end{equation*}

\noindent
Finally, we use the RMSProp \cite{46} optimizer with a decay of 0.99 and an epsilon of 0.1. The network of the VMA3C method includes 4 layers: 2 convolutional layers (one has 16 filters of 8 $\times$ 8 with a stride of 4 followed by another one that has 32 filters of 4 $\times$ 4 with a stride of 2), one fully-connected layer with a size of 256 and ReLU activation functions, and one output layer that includes a single critic and and set of $N$ actors ($N=2$ for two-agent setting and $N=3$ for three-agent setting). The critic is connected into a linear activation function while each actor is connected into a \emph{softmax} activation function.

\begin{figure}[!t]
\centering
\includegraphics[width=1.0\linewidth]{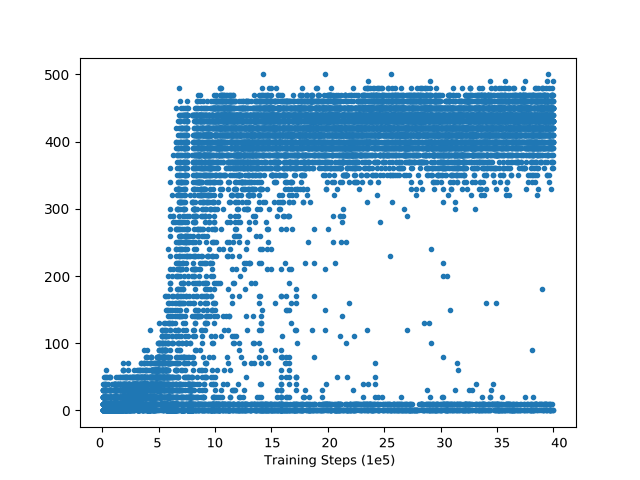}
\caption{The reward distribution of VMA3C in the two-agent setting of Milk Factory.}
\label{fig:8} 
\end{figure}

We use A3C and VMA3C to train agents in 4 million steps. It takes approximately 12 training hours for each variant. During the training process, we create 40 checkpoints in different training steps. For each checkpoint, we perform a 10,000-step evaluation and record the mean of total reward and its standard error. We divide the performance evaluation into three stages. Firstly, we compare the performance of VMA3C with A3C in the two-agent setting of Milk Factory (one pickup robot and one mechanic robot). Subsequently, we add an additional pick-up robot to examine the scalability of the proposed scheme. Finally, we re-evaluate the previous two stages with different values of \emph{ER} to examine the robustness of VMA3C in stochastic environments.

\subsection{Two-agent setting}
\label{sec:4.1}

In this setting, two robots are operated in Milk Factory: one pick-up robot and one mechanic robot. We examine the cooperation between the robots to maximize the total reward. Because the maximum number of episode steps is set to 200, it is feasible to calculate the optimal score that the two robots can achieve. In an ideal case, it takes 8 steps to pick a milk bottle, go to the box, and put the bottle into the box. The robot optimally achieves a reward of 20 in each round. Therefore, the maximum reward that the robot can achieve is $200*20/8=500$ without any errors. If an error occurs during the operation, the total reward must be smaller because the robot misses the current milk on the conveyor belt. Therefore, the reward of 500 is also the optimal total reward of two robots. Fig. \ref{fig:8} describes the reward distribution of VMA3C during the training. We can see that VMA3C takes only three training hours to establish an optimal policy that approximately reaches a reward of 500. It is possible due to the visual map aids the mechanic robot to fix the pick-up robot immediately when an error occurs.

\begin{figure}[!h]
\centering
\includegraphics[width=1.0\linewidth]{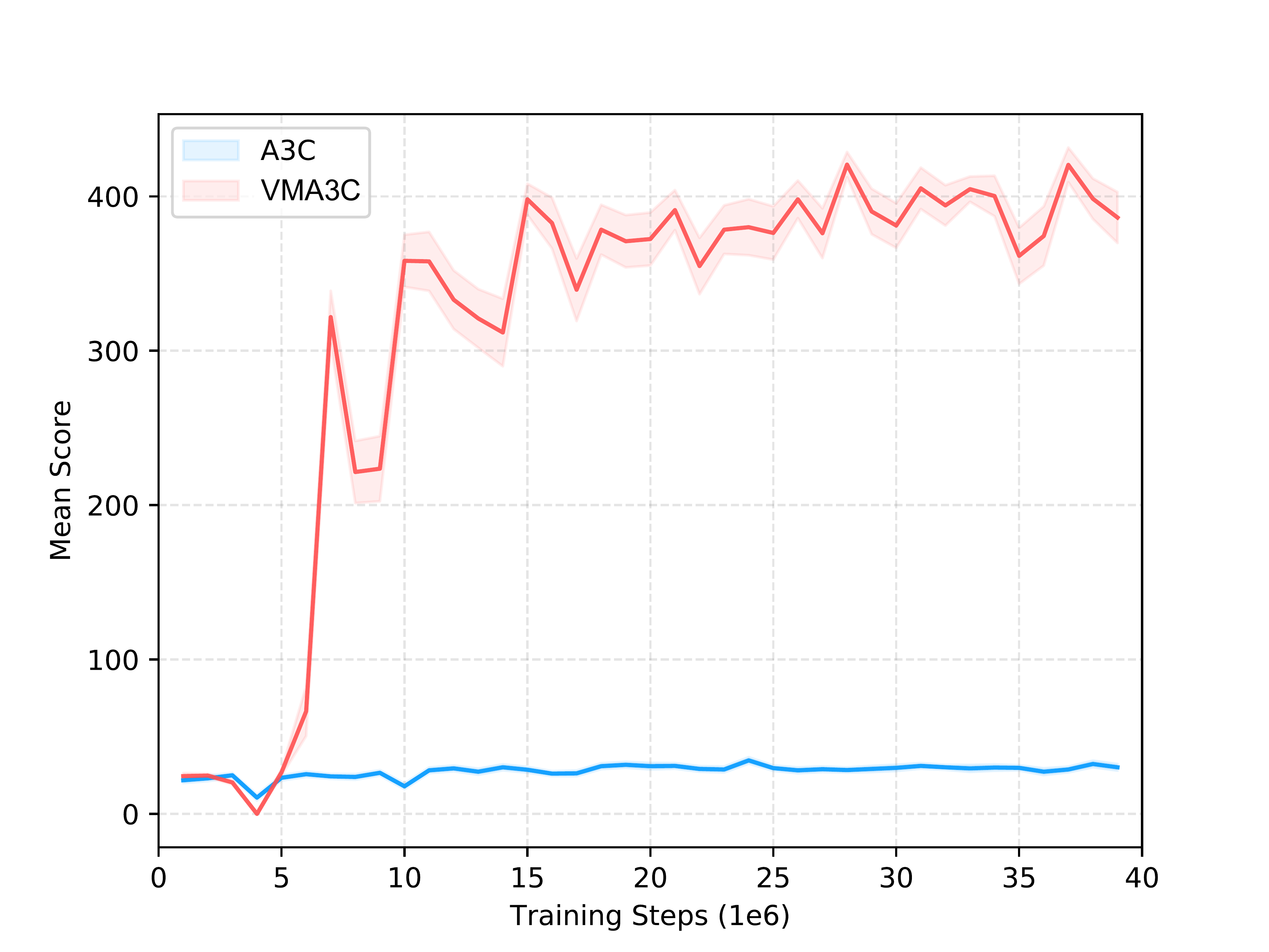}
\caption{Comparison of the mean of total rewards between two methods (A3C and VMA3C) in the two-agent setting of Milk Factory.}
\label{fig:9} 
\end{figure}

Fig. \ref{fig:9} shows the performance of two methods, A3C and VMA3C, regarding the mean of total rewards. In the figure, A3C performs poorly without any improvements after 12 hours of training. Furthermore, the pick-up robot gets stuck after picking a milk bottle. The mechanic robot is located next to it, as shown in the following video \emph{https://youtu.be/J0qusnfyrr0}. As opposed to A3C, VMA3C achieves an optimal policy after 3 hours of training. This marks the importance of visual communication map in multi-agent problems. The pick-up robot in VMA3C is operated as expected: it picks a milk bottle, goes to the box by the shortest path, puts the milk bottle into the box, and goes back to the conveyor belt. Moreover, the mechanic robot fixes the pick-up robot if necessary.

\subsection{Three-agent setting}
\label{sec:4.2}

\begin{figure}[!h]
\centering
\includegraphics[width=1.0\linewidth]{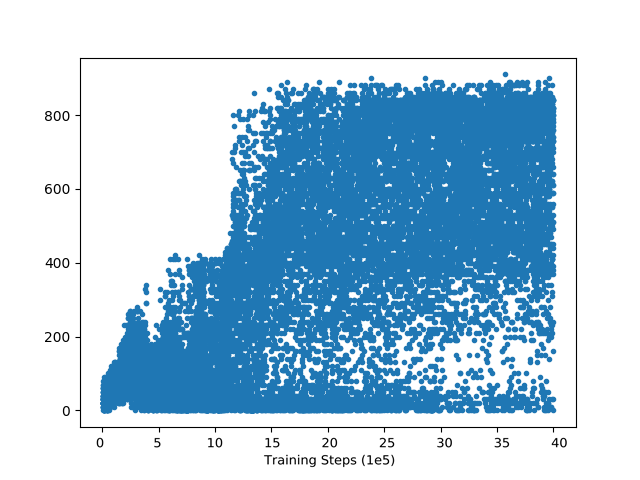}
\caption{The reward distribution of VMA3C in the three-agent setting of Milk Factory.}
\label{fig:10} 
\end{figure}

To examine the scalability of VMA3C, we add a pick-up robot into the gameplay. Therefore, we have two pick-up robots and one mechanic robot. We also add an additional box and a milk bottle on the conveyor to increase the productivity. Fig. \ref{fig:10} shows the reward distribution of the VMA3C method in the three-agent setting. It can estimate an optimal policy that reaches a reward of 900. However, the results vary wildly because it is impossible to fix two pick-up robots at the same time. In this case, the environment is more stochastic than the previous one. 

\begin{figure}[!t]
\centering
\includegraphics[width=1.0\linewidth]{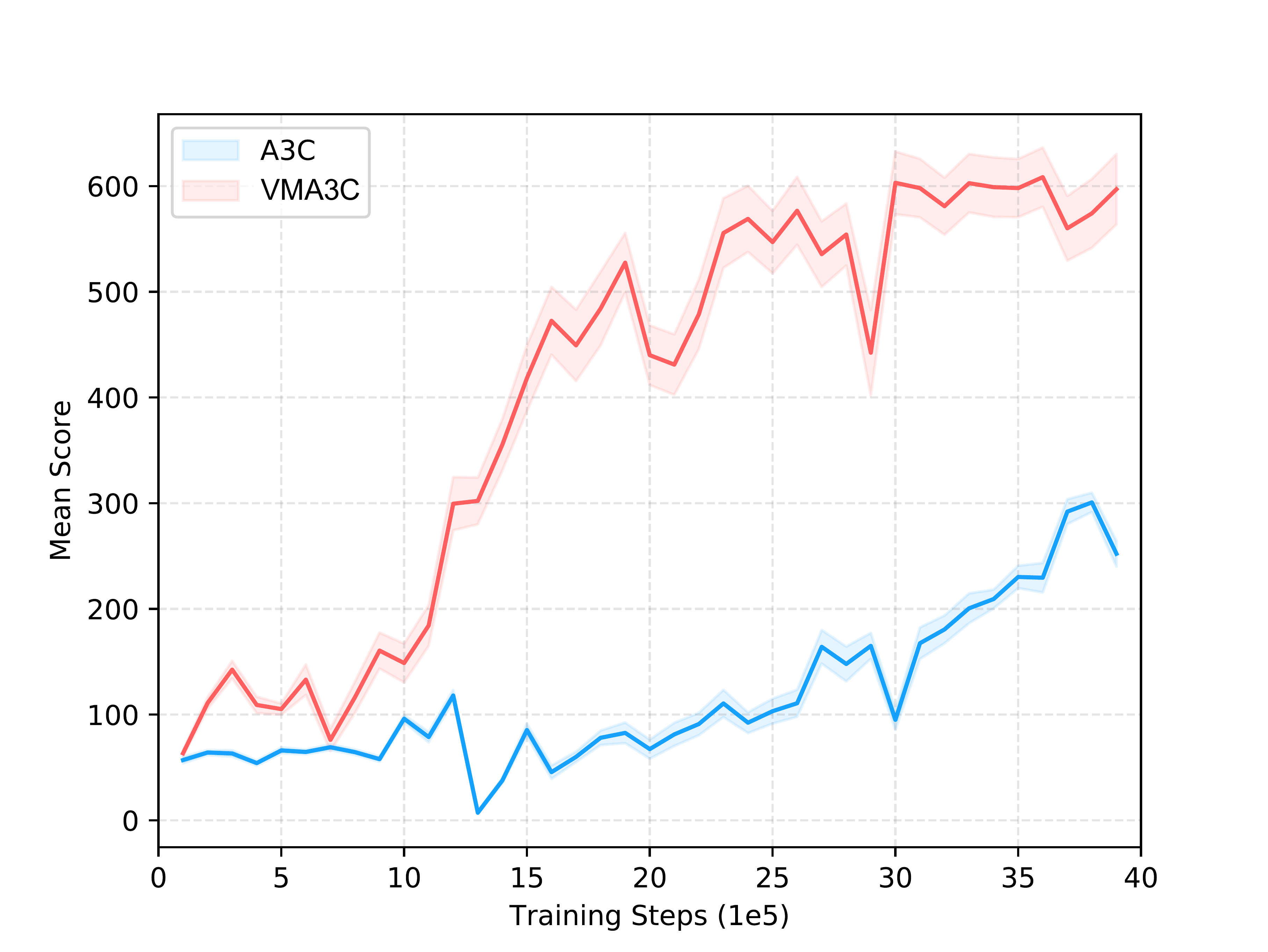}
\caption{Comparison of the mean of total rewards between two methods (A3C and VMA3C) in the three-agent setting of Milk Factory.}
\label{fig:11} 
\end{figure}

Fig. \ref{fig:11} compares the performance of VMA3C and A3C in the three-agent setting of Milk Factory. The A3C method obtains a maximum reward of 300 after 12 hours of training while the VMA3C has a 200\% higher performance than the A3C method. As shown in the following video, two pick-up robots in VMA3C operate concurrently while the mechanic robot is located in the middle of the screen, as shown in the following video \emph{https://youtu.be/eoud2D0nW1k}.

\subsection{Robustness}

\begin{figure*}[!t]
\includegraphics[width=0.5\textwidth]{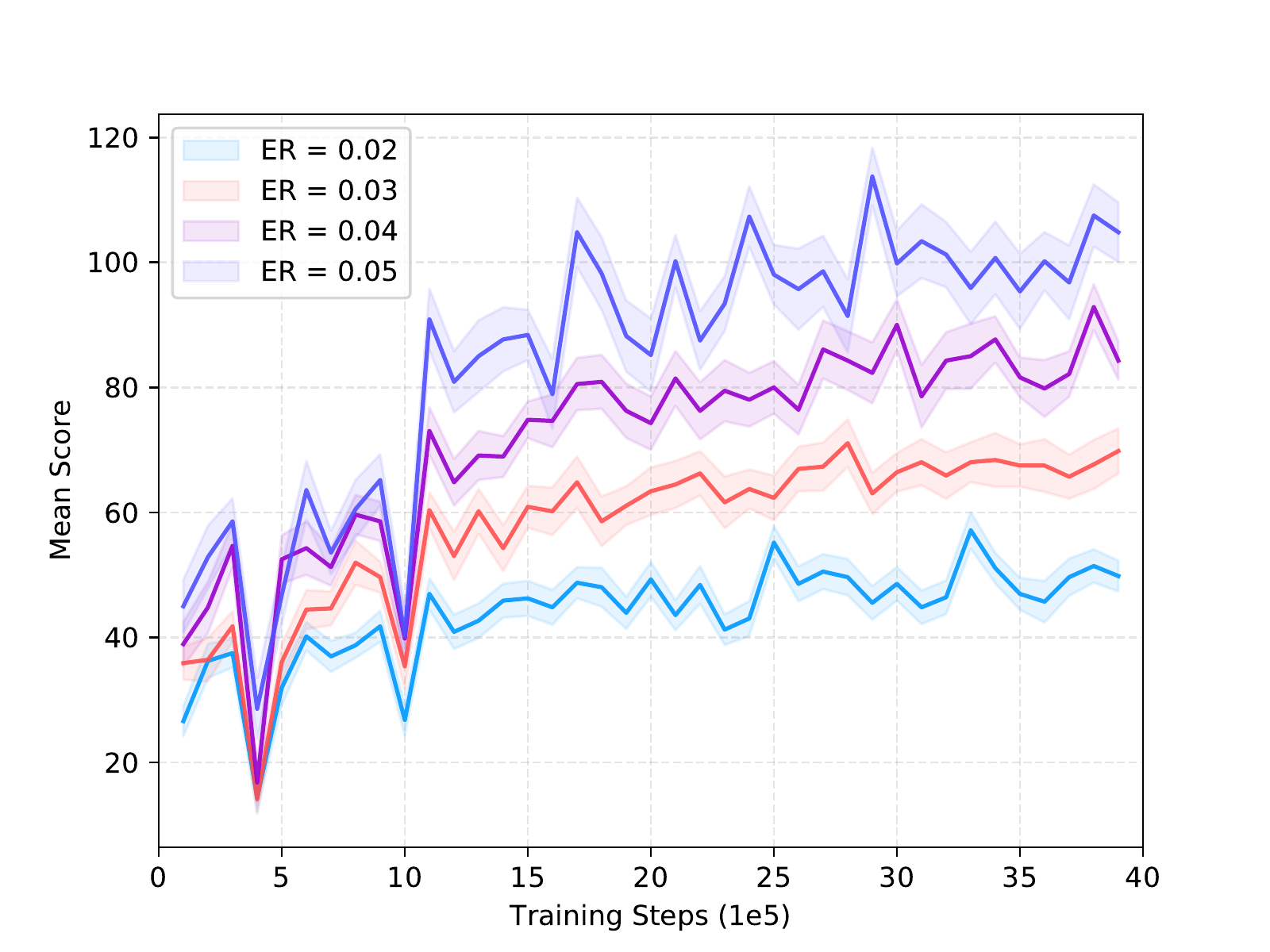}
\includegraphics[width=0.5\textwidth]{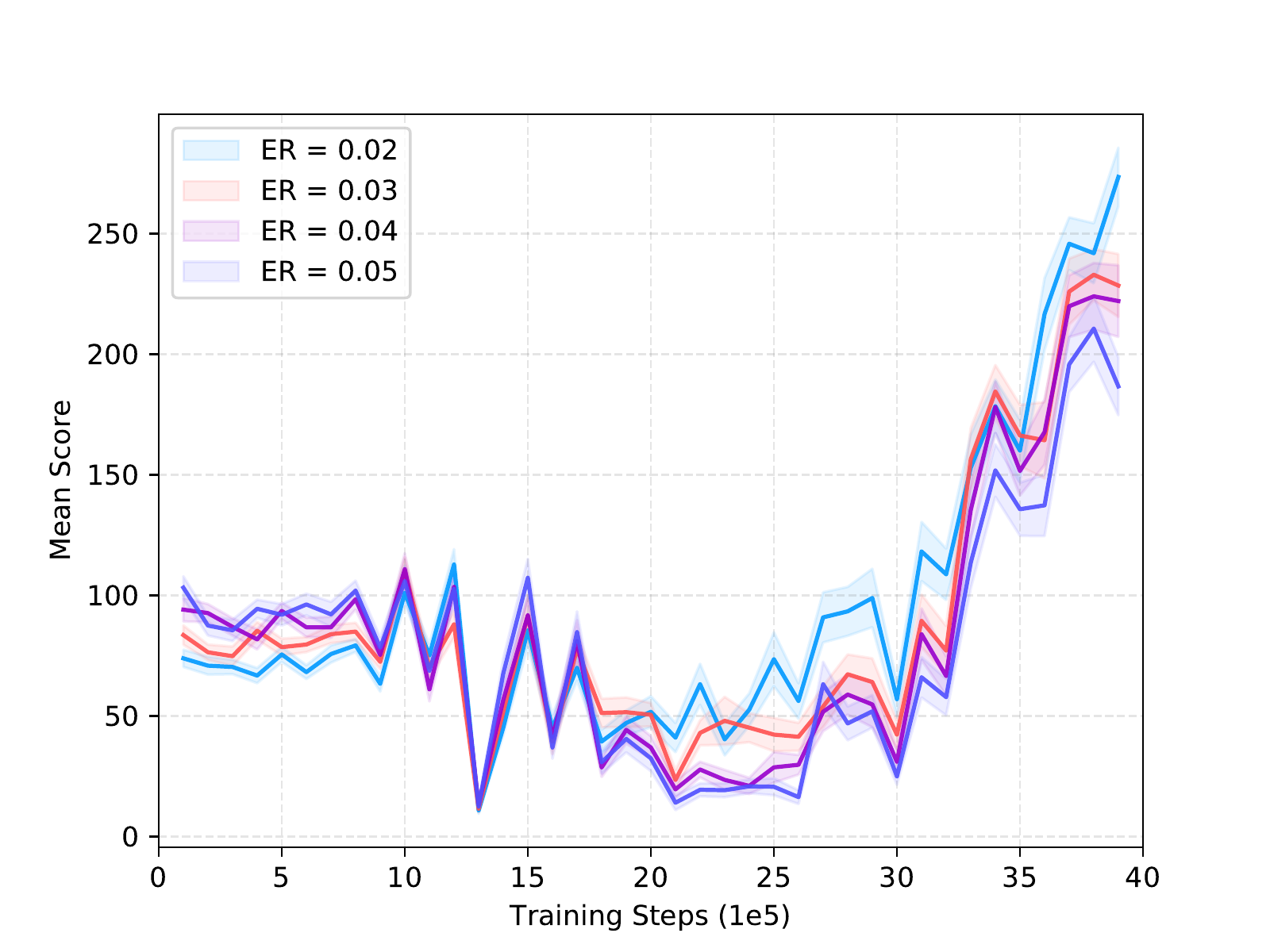}
\caption{The performance of A3C in Milk Factory with four different error rates: $2\%$, $3\%$, $4\%$, and $5\%$. The left figure is conducted in the two-agent setting and the right figure is conducted in the three-agent setting of Milk Factory. }
\label{fig:12} 
\end{figure*}

\begin{figure*}[!t]
\includegraphics[width=0.5\textwidth]{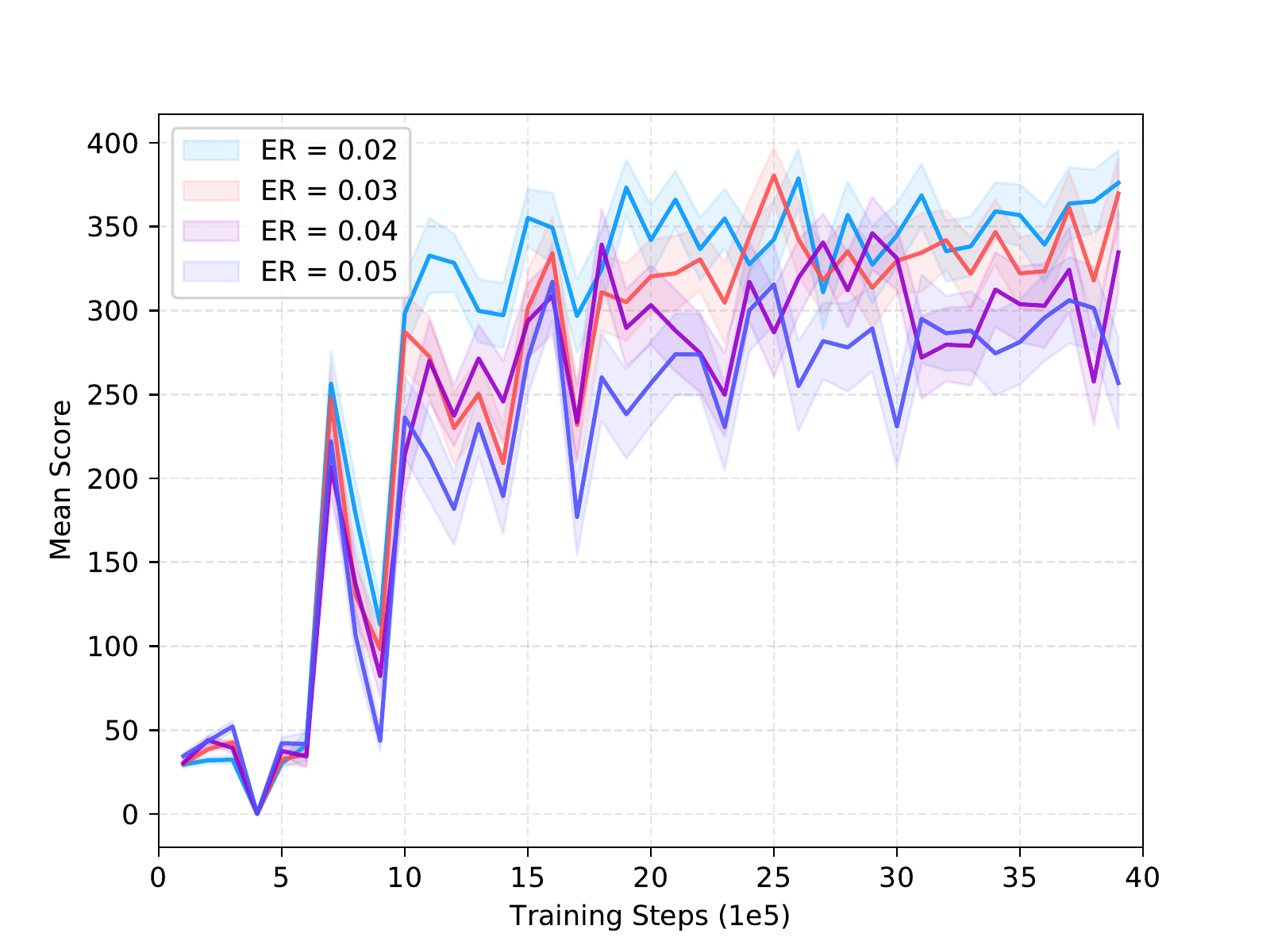}
\includegraphics[width=0.5\textwidth]{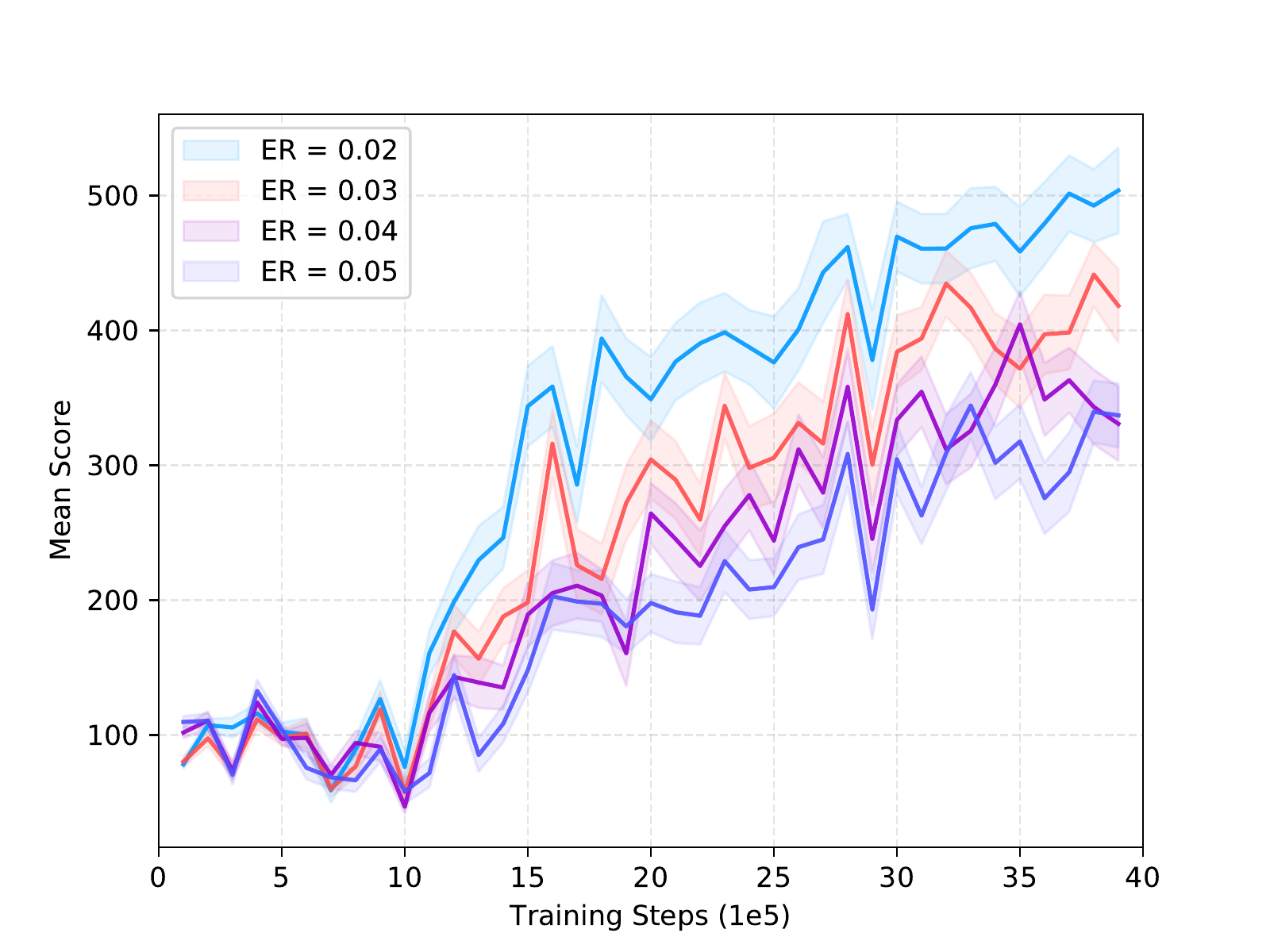}
\caption{The performance of VMA3C in Milk Factory with four different error rates: $2\%$, $3\%$, $4\%$, and $5\%$. The left figure is conducted in the two-agent setting and the right figure is conducted in the three-agent setting of Milk Factory. }
\label{fig:13} 
\end{figure*}

In this subsection, we vary the values of \emph{ER} to examine the robustness of the proposed method. We record the performance of both approaches by using four different error rates: $2\%$, $3\%$, $4\%$, and $5\%$. Fig. \ref{fig:12} shows the performance of A3C in the two-agent setting and the three-agent setting of Milk Factory. In the two-agent setting, A3C performs poorly, as explained in the previous subsection. We notice that the mean of total reward increases with the error rate. This implies that the total reward is calculated by the mechanic robot. Finally, Fig. \ref{fig:13} presents the performance of VMA3C in the two and three-agent setting of Milk Factory. We conclude that VMA3C is robust in the stochastic environment because it still achieves a high reward regardless of the error rate.

\section{Conclusions}
\label{sec:5}
Communications between multiple agents are difficult to implement, especially when agents are characterized by deep RL algorithms. The paper conducts an interesting investigation of using virtual communication channel via visual indicators to establish a cooperative policy for multi-agent problems. The method has practical meaning and can be applied widely in real-world applications due to many beneficial factors. Firstly, it can be used with any standard RL methods. Secondly, the proposed method is scalable in terms of the number of agents and the type of agents. Finally, the method is robust in stochastic environments and plays a vital role to solve the non-stationary problem in multi-agent domains. By using visual communication map, the agents learn a cooperative policy on their own to maximize the reward in a short period of training time. We continue to work on the effect of visual indicators to multi-agent domains in different directions:

\begin{itemize}
\item The method requires human knowledge to reduce the number of visual indicators, which prevents the automation of the approach. We suggest using a Bayesian model to predict the dependency between agents to completely automate the proposed method.

\item The scheme is constructed from the A3C method and evaluated in a discrete action environment. We will extend the use of visual communication map with different single-agent deep RL algorithms and examine it in continuous action domains.

\end{itemize} 

\section*{Acknowledgement}

The authors wish to thank our colleagues in Institute for Intelligent Systems Research and Innovation for their comments and helpful discussions.




\end{document}